%% file: main.tex
\documentclass[letterpaper]{article} 
\usepackage{aaai2026}  
\usepackage{times}  
\usepackage{helvet}  
\usepackage{courier}  
\usepackage[hyphens]{url}  
\usepackage{graphicx} 
\usepackage{natbib}  
\usepackage{caption} 
\usepackage{algorithm}

\usepackage{newfloat}
\usepackage{listings}
\DeclareCaptionStyle{ruled}{labelfont=normalfont,labelsep=colon,strut=off} 
\floatstyle{ruled}
\newfloat{listing}{tb}{lst}{}
\floatname{listing}{Listing}
\usepackage{xcolor}
\usepackage{latexsym}
\usepackage{multirow}
\usepackage{siunitx}
\usepackage[most]{tcolorbox}

\usepackage[utf8]{inputenc}

\usepackage{microtype}

\usepackage{inconsolata}

\usepackage{booktabs} 
\usepackage{amsmath}
\usepackage{mathtools}
\usepackage{multirow}
\usepackage{siunitx}
\usepackage{lineno}
\usepackage{enumitem}
\usepackage{amsfonts}       
\usepackage{nicefrac}       
\usepackage{amsmath}
\usepackage{amssymb}
\usepackage[noend]{algpseudocode}
\usepackage{subcaption}
\usepackage[textsize=scriptsize]{todonotes}
\usepackage{cuted}

\usepackage{array} 

\input{new-commands}

\title{
PRECISE: Reducing the Bias of LLM Evaluations Using Prediction-Powered Ranking Estimation
}
\author {
    Abhishek Divekar\thanks{Primary contributor and corresponding author.}, Anirban Majumder
}
\affiliations {
    Amazon AI\\
    adivekar@amazon.com, majumda@amazon.com
}

\begin{document}

\maketitle


\input{sections/0_Abstract}
\input{sections/1a_Introduction}
\input{sections/1b_Application}
\input{sections/2_Background}
\input{sections/3_Method}
\input{sections/4a_Datasets}
\input{sections/4b_Results}
\input{sections/4c_Results_Application}

\input{sections/5_Conclusion}
\input{sections/6_Future_work}

\appendix
\input{appendices/calibration}
\input{appendices/query_prompts}

\section{Acknowledgments}
Financial support for experiments was provided by Amazon Central Machine Learning department. We additionally thank Suhas Kowshik for providing feedback on the methodological framing.

\bibliography{aaai2026}

\end{document}

%% file: new-commands.tex
\definecolor{todoblue}{RGB}{33,150,243} 

\definecolor{pink-color}{RGB}{255,1,97} 

\definecolor{red-color}{RGB}{255,111,105} 
\definecolor{dark-red-color}{RGB}{174,0,1} 
\definecolor{light-red-color}{RGB}{255,152,150}

\definecolor{green-color}{RGB}{136,216,176} 
\definecolor{dark-green-color}{RGB}{49,163,84} 
\definecolor{light-green-color}{RGB}{136,216,176} 
\definecolor{sea-green-color}{RGB}{95,158,160} 

\definecolor{dark-blue-color}{RGB}{49,130,189} 
\definecolor{light-blue-color}{RGB}{158,202,225} 

\definecolor{grey-color}{RGB}{167,173,186}
\definecolor{dark-grey-color}{RGB}{79,91,102}
\definecolor{gold-color}{RGB}{247,205,15} 

\newcommand{\aref}[1]{\hyperref[#1]{Appendix~\ref*{#1}}}

\newcommand{\lowerbetter}{\ensuremath{(\downarrow)}}

\newcommand{\ignore}[1]{}
\newcommand{\ig}[1]{}

\newtcolorbox[list inside=prompt]{prompt}[1][]{
    colbacktitle=black!60,
    coltitle=white,
    fontupper=\footnotesize,
    boxsep=5pt,
    left=0pt,
    right=0pt,
    top=0pt,
    bottom=0pt,
    boxrule=1pt,
    title={Prompt~\thetcbcounter}, 
    #1,
}

\newcommand{\urlsmall}[1]{{\scriptsize{\url{\detokenize{#1}}}}}

\newcolumntype{h}{>{\setbox0=\hbox\bgroup}c<{\egroup}@{}}
\newcolumntype{C}[1]{>{\centering\arraybackslash}m{#1}}
\newcolumntype{R}[1]{>{\raggedleft\arraybackslash}p{#1}}
\newcolumntype{L}[1]{>{\raggedright\arraybackslash}p{#1}}

\newcommand{\distas}[1]{\mathbin{\overset{#1}{\kern0pt\sim}}}%

\def\ours{{\textsc{PRECISE}}}

%% file: sections/0_Abstract.tex
\begin{abstract}
Evaluating the quality of search systems traditionally requires a significant number of human relevance annotations. In recent times, several systems have explored the usage of Large Language Models (LLMs) as automated judges for this task while their inherent biases prevent direct use for metric estimation. We present a statistical framework extending Prediction-Powered Inference (PPI)~\cite{angelopoulos2024ppiefficientpredictionpoweredinference} that combines minimal human annotations with LLM judgments to produce reliable estimates of metrics which require sub-instance annotations. Our method requires as few as $100$ human-annotated queries and $10,000$ unlabeled examples, reducing annotation requirements significantly compared to traditional approaches. We formulate our proposed framework (\ours) for inference of relevance uplift for an LLM-based query reformulation application, extending PPI to sub-instance annotations at the query-document level. By reformulating the metric-integration space, we reduced the computational complexity from $O(2^{|C|})$ to $O(2^K)$, where $|C|$ represents corpus size (in order of millions). Detailed experiments across prominent retrieval datasets demonstrate that our method reduces the variance of estimates for the business-critical Precision@K metric, while effectively correcting for LLM bias in low-resource settings.
\end{abstract}

%% file: sections/1a_Introduction.tex
\input{figure/application}

\section{Introduction}
Large Language Models (LLMs)~\cite{Achiam2023GPT4TRShort, bai2022constitutionalaiharmlessnessaiShort, deepseekai2025deepseekv3technicalreportShort} have rapidly gained traction in industrial applications.
Evaluation of LLM applications traditionally relies on human audits, a process that is neither scalable nor cost-effective, especially when dealing with large, diverse datasets collected from real-world applications. To address this challenge, recent work~\cite{saad-falcon-etal-2024-ares, 10.5555/3666122.3668142, es-etal-2024-ragas, dong-etal-2024-llm} has explored using LLMs themselves as evaluators, leveraging their strong reasoning capabilities and contextual comprehension. This offers a potential solution to the evaluation bottleneck, automating quality assessment of complex tasks at scale.

Ranking and recommendation problems are cornerstones of today's e-commerce websites, spanning search, advertising, and product recommendations.
Human evaluation has traditionally been the gold standard for evaluating ranking quality; however, it faces unique challenges in this domain. Ranking models and algorithms change frequently, necessitating repeated evaluations. Relying on implicit signals like user clicks for evaluation can introduce biases~\cite{10.1145/3366423.3380255, 10.1145/2911451.2911537}, as clicks are influenced by factors other than relevance, such as position and presentation.

\input{figure/high_level_diagram}

LLM-based evaluation has thus emerged as a promising alternative, potentially enabling efficient and timely assessment of large-scale recommendations or search results. 

However, this is not without risks, including potential biases inherent in LLMs~\cite{chen-etal-2024-humans, li-etal-2024-split} and consistency issues across different contexts~\cite{shen-etal-2023-large}. These challenges necessitate careful consideration and mitigation strategies when leveraging LLMs to evaluate e-commerce ranking and recommendation algorithms.

While human evaluation is crucial for unbiased assessment of ranking and recommendation systems, it is limited in quantity due to cost and scale challenges. Conversely, LLM evaluations are abundant but potentially biased. 
To leverage the strengths of both human and LLM evaluations, we formulate a novel ranking-metric estimator based on Prediction-Powered Inference (PPI)~\cite{doi:10.1126/science.adi6000}. PPI is a framework for valid statistical estimation, where limited human annotations are augmented with machine learning predictions. Ranking systems present a unique challenge for PPI, due to the hierarchical nature of the estimation task: while human annotations are collected at the atomic (query, document) level, ranking performance metrics are computed at query level and then aggregated over the entire dataset. This inconsistency makes the vanilla PPI estimator infeasible. 

We address this gap by extending the PPI framework to estimate from signals provided from sub-query level human and LLM annotations, demonstrating that our technique is compatible with standard ranking metrics such as Precision@K. Our comprehensive evaluation across proprietary and public datasets demonstrates the framework's effectiveness across diverse e-commerce search systems and multiple evaluator models. 

%% file: figure/application.tex
\begin{figure}[!t]
\centering
\hspace{-1.5ex} 
\includegraphics[width=0.45\textwidth]{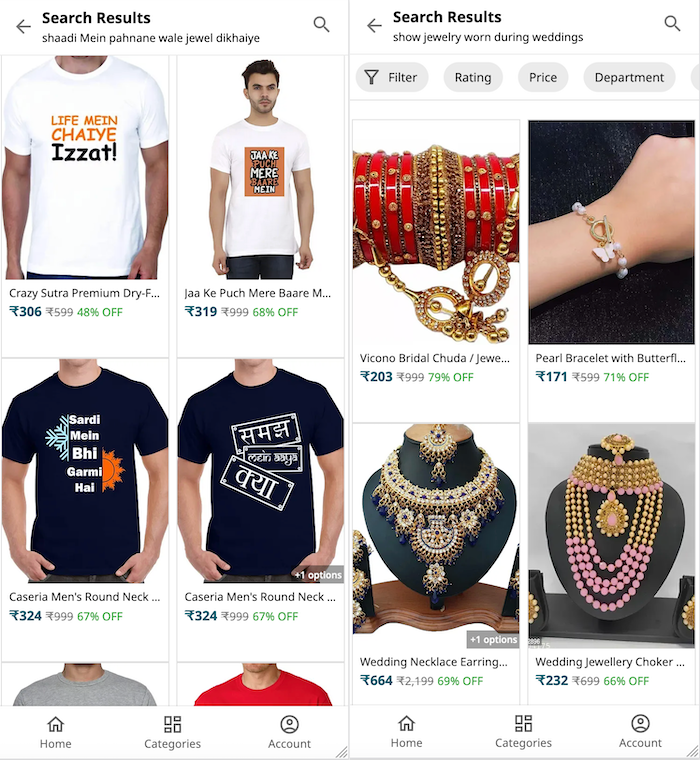}
\caption{
    Code-mixed queries encountered in our production system, demonstrating the linguistic challenges of Indian e-commerce search. Left: queries from customers often mix Hindi words written in Latin script with English. Right: query-reformulation into grammatical English using a frontier LLM greatly improves search relevance. Our deployed approach \ours-PPI seeks to estimate the performance of the query-reformulation approach by debiasing LLM relevance judgements with minimal human annotations.
}
\label{fig:hinglish}
\end{figure}

%% file: figure/high_level_diagram.tex
\begin{figure*}[t!]
    \centering
    \includegraphics[width=0.95\textwidth]{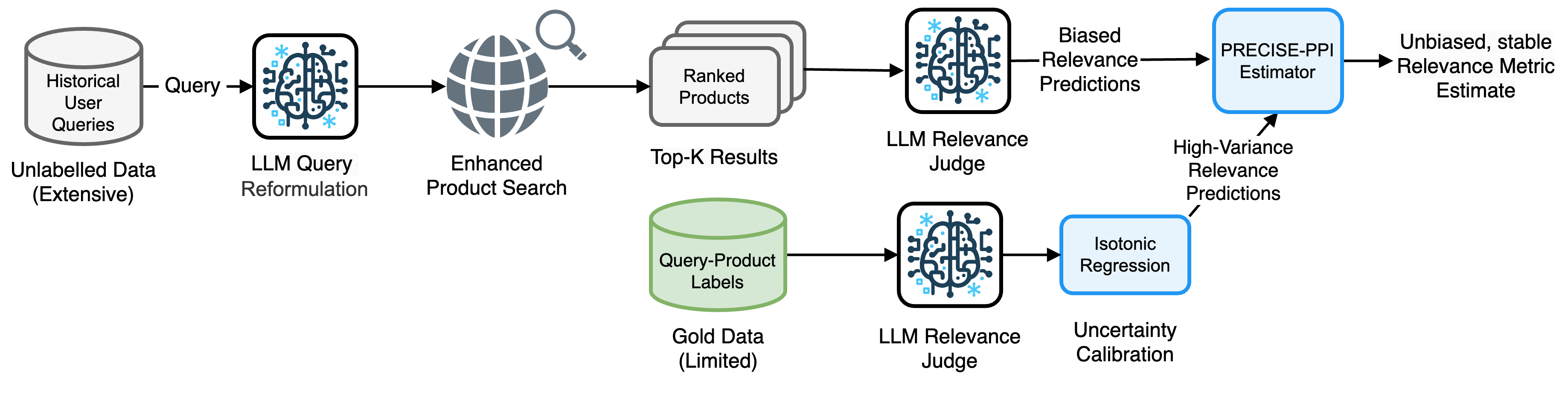}
    \caption{
    High-level flow of our \ours-PPI method to estimate relevance metrics. Our approach combines estimates from LLM annotations on unlabelled queries and human-labelled gold annotations of query-product relevance.
    }
    \label{fig:high-level-digram}
\end{figure*}

%% file: sections/1b_Application.tex
\section{Application Description}
\label{sec:application}

We consider an e-commerce application scenario in India that enables resellers to purchase products on behalf of end customers who have limited proficiency with traditional e-commerce websites or mobile apps, particularly in Tier-2 and Tier-3 Indian cities. The application maintains an extensive catalog featuring millions of buyable products, while leveraging Amazon fulfillment to ensure reliable delivery and returns. The application focuses primarily on low-value products in the fashion and electronics categories.

This application lacks a dedicated set of annotators to provide a large set of human-curated relevance judgments. This posed a unique challenge to estimate the quality of new search relevance improvements. In the rest of this paper, we analyze how our estimation approach was used to guide the deployment of system that uses LLM-based query reformulation to overcome significant linguistic challenges that majorly impact search performance.

\subsection{Query Pattern Analysis}

We categorized queries into three volume-based segments: Head queries (contributing 50\% of total search volume), Body queries (25\% of volume), and Tail queries (25\% of volume). We conducted a comprehensive analysis of 1,000 queries from each segment using LLM-as-a-Judge, revealing substantial linguistic defects that impair the effectiveness of the production search engine:

\begin{itemize}
\item \textbf{Increasing defect severity}: The fraction of problematic queries increases dramatically from Head to Tail segments, with organic grammatically correct English queries decreasing sharply from Head queries to Tail queries.

\item \textbf{Hinglish prevalence}: Hinglish queries (Hindi words written in Latin script) represent a significant portion of search volume, particularly in Body and Tail segments. Figure~\ref{fig:hinglish} illustrates typical examples of such queries.
\end{itemize}

With millions of unique queries across all segments, these linguistic defects significantly affect daily customer search quality.

\subsection{LLM-Based Query Reformulation Solution}

To address these query defects, we developed an LLM-based query reformulation system using Claude 3 Sonnet. The system employs two specialized prompts which use both reasoning traces and in-context exemplars:

\begin{itemize}
\item \textbf{Prompt V1}: Performs translation of Hinglish queries and correction of grammatical errors and typos in customer-entered queries.

\item \textbf{Prompt V2}: Extends V1 with Indian ethnic context awareness to preserve culturally-specific terms (e.g. ``kurti'', ``salwar kameez'') that should not be translated, as these terms appear as-is in product catalogs.
\end{itemize}

The reformulation system targets all Head and Body queries, covering 75\% of total search volume. Tail queries are excluded due to their high uniqueness (the vast majority are searched only once) and the prohibitive cost of reformulating several million queries within the launch timeline of the Diwali sale. 

Note that we anonymize the user queries to remove potential PII information prior to usage in our LLM-based reformulation solution.

\subsection{Metric Estimation Pre-Deployment}

Query reformulation presents a fundamental challenge: it can either significantly improve or severely degrade search relevance, depending on the quality of the reformulations. In traditional search experiments, the impact of ML solutions is validated using extensive human-annotated test sets of relevance judgments. However, our application's annotation constraints made this approach infeasible.

Deploying an untested query reformulation system would pose substantial business risk. The available audit bandwidth consisted of only a few days of software engineering team time immediately before launch. This scenario exemplifies the exact use case for which our \ours{} approach was designed: estimating the true performance impact of an ML system when extensive human annotation is prohibitively expensive or time-constrained, but where deployment decisions must be made with confidence.

We deployed \ours{} to estimate Precision@K improvements across three treatments: (a) Control: unmodified production queries; (b) T1: queries reformulated with Prompt V1; (c) T2: queries reformulated with Prompt V2 including Indian ethnic context. Our framework correctly identified the best-performing treatment, which was subsequently validated through A/B testing with limited traffic and deployed to production, improving search relevance for millions of users and leading to significant business impact for our application.

%% file: sections/2_Background.tex
\section{Method}
\label{sec:method}
In this section, we introduce our novel \ours{} method of evaluating ranking models using LLMs. We first describe the general framework of Prediction-Powered Inference for estimating performance metrics \cite{boyeau2025autoeval}. 

\subsection{Background: PPI for Metric Estimation}

Assume we have a human-labeled ``gold'' dataset \mbox{$\mathcal{D}_g=\{(x_g^{(1)}, y_g^{(1)}) \cdots (x_g^{(n)}, y_g^{(n)})\}$} and have access to an unlabeled dataset $\mathcal{D}_u=\{x_u^{(1)}, \cdots, x_u^{(N)}\}$ where $N \gg n$, and both covariates are iid samples from the same (true) distribution. 
Our goal is to evaluate performance of a machine learning system $f$ using the datasets $\mathcal{D}_g$ and $\mathcal{D}_u$. Let $\phi$ be any metric of interest e.g. accuracy for classification task, squared error for regression etc. 
We can estimate model performance as the expectation of $ \phi(f(x_g^{(i)}), y_g^{(i)})$ over the labelled data; however the same cannot be done with $\mathcal{D}_u$, due to absence of ground-truth labels. Since we have limited labeled examples, reporting $f$ on $\mathcal{D}_g$ exhibits high variance in the accuracy estimate.

To leverage the  large corpus of unlabeled data, we can employ an ``annotator'' ML model $M$ that generates synthetic labels $\{\tilde{y}_{u}^{(1)}, \cdots, \tilde{y}_{u}^{(N)} \}$ and average 
$\phi(f(x_u^{(i)}), \tilde{y}_u^{(i)})$ across $\mathcal{D}_u$. While this reduces variance, potential bias from the trained model $M$ can creep in, resulting in a statistically biased estimate. Prediction-Powered Inference is a statistical framework to debias estimates by leveraging both labeled and unlabeled datasets. We typically use the efficient PPI++ estimator \cite{angelopoulos2024ppiefficientpredictionpoweredinference}:

\begin{align}
\notag
&\hat{\mu}_{\textsc{PPI}++} = \lambda \left[ \frac{1}{N} \sum_{i=1}^{N} \tilde{\mu}_u^{(i)} \right]\\
&+ \frac{1}{n} \sum_{i=1}^{n} \left[ \phi(f(x_g^{(i)}), y_g^{(i)}) - \lambda \cdot \tilde{\mu}_g^{(i)} \right]
\end{align}

where,

\begin{align}
\tilde{\mu}_u^{(i)} = \underset{y \sim M(\cdot \mid x_u^{(i)})}{ \mathbb{E}}{} \phi(f(x_u^{(i)}), y)
\end{align}

is the estimate of the metric on each instance of the unlabelled set, using the conditional probability distribution output from the annotator over the output space $Y$ as in \cite{boyeau2025autoeval}. 
Each $\tilde{\mu}_g^{(i)}$ is calculated analogously. 

Here $0 \le \lambda \le 1$ is a hyperparameter that can be tuned to minimize the variance of the estimator $\mu_{\textsc{PPI}++}$. However, the estimator remains unbiased for any value of $\lambda > 0$.

%% file: sections/3_Method.tex
\subsection{\ours-PPI: Ranking Metric Estimation}

A limitation of the previous PPI formulation is that it is undefined for situations where the annotator model provides synthetic labels at a granularity other than the instance-level. For example, in the case of estimating common ranking metrics such as Precision@K, Recall@K, etc, the notion of an ``instance'' pertains to a query but the annotator model provides a relevance annotation at the query-document level.

The key challenge here is the formulation of the output space $y \in Y$ over which to take the integrand/summand $\phi(f(x^{(i)}), y) \cdot \tilde{p}^{(i)}(y)$, which is also compatible with the granularity of $\tilde{p}^{(i)}(y) = M(y | x^{(i)})$ provided by the annotator model. 

To overcome this issue, we reformulate $\tilde{\mu}_u^{(i)}$ and $\tilde{\mu}_g^{(i)}$ in order to estimate $\hat{\mu}_{\textsc{ppi}++}$ appropriately for the task of search relevance. Concretely, assume the corpus of documents $C = \{d^{(1)},\ldots,d^{(|C|)}\}$ is an internal aspect of the search relevance model under evaluation, i.e. $f(x) = f_C(x)$, where $x$ is a single query. Assume this model provides binarized relevance labels to $K$ documents in $C$. We can imagine the prediction as a K-hot vector:

\[\hat{y} = f_C(x) = \left[rel(d^{(1)}),\ldots,rel(d^{(|C|)})\right],\]

where $\|\hat{y}\|_1 = K$.  An example realization may be $[1,0,1,\ldots,0]$; exactly K indexes must be hot.

Assume that for the purpose of estimating Precision@K using PPI, we have labelled a small dataset of $n$ queries, providing a binary relevance annotation to each of the top-K results per query. In this scenario, we can represent the ground-truths for the gold set as using a similar one-hot vector:

\[y = \left[rel(d^{(1)}),\ldots,rel(d^{(|C|)})\right],\]

where $\|y\|_1 \leq K$ and \textit{at most} K values are ``hot''.

To measure Precision@K at the instance-level, we would simply calculate the scaled dot product of these quantities:

\[\phi(f_C(x), y) = \phi(\hat{y},y) = \frac{\hat{y}^Ty}{K}\]

However, both $y$ and $\hat{y}$ are sparse; it is equivalent to compute the dot product of the K documents which are marked as relevant by $f_C(\cdot)$.

The above observation is crucial to the efficient formulation of the iterable space $Y$ which we integrate/sum to produce $\tilde{\mu}_u^{(i)}$ and $\tilde{\mu}_g^{(i)}$. An exact calculation of these quantities would require us to consider $Y$ to be all vectors of length $|C|$, and considering every possible combination of hot values, i.e. $Y = \{0,1\}^{|C|}$. As $|C|$ is often in millions, this calculation is intractable.

However, due to the sparsity in the calculation of Precision@K (as we have at most K ``hot'' positions), we can instead iterate over a much-reduced space of all combinations of K-length vectors $Y = \{0,1\}^K$. 

Our key observation here is that the probability mass of all $|C|$-length vectors where the K documents are zeros, is accumulated into a single probability weight of the all-zero K-length vector. This makes the computation tractable: although the size of the iterable space $|Y|$ is still exponential, typically we estimate Precision@K with small $K$ (e.g. $\leq 10$), permitting us to estimate $\tilde{\mu}_u^{(i)}$ and $\tilde{\mu}_g^{(i)}$.

Concretely, consider a single query $x$ for which we have a K-length vector of annotator-provided probabilities \mbox{$\tilde{p}'(d_k) = M(d_k | x)$} that the $k$th ranked document $d_k$ is relevant to the query $x$. 

We can convert this into a probability value for each K-length binary vector $y \in Y = \{0,1\}^K$ by applying the probability-distribution operation:

\begin{align}
\tilde{p}(y) = \prod_{k=1}^K \tilde{p}'(d_k)^{y_k}(1-\tilde{p}'(d_k))^{(1-y_k)}
\end{align}

where $Y = \{0,1\}^K$ is all possible K-length binary vectors and $y_k$ is each element of $y \in Y$.




The calculation of $\hat{\mu}_{\textsc{ppi}++}$ then proceeds as before for regular PPI. Thus, we are able to formulate the estimate for both ranking and information retrieval tasks. 

%% file: sections/4a_Datasets.tex
\input{figure/esci_precision_results}

\section{Experimental Setup}
\label{sec:expt-setup}

\subsection{Datasets}

We conduct experiments on two complementary datasets to validate \ours{} for search relevance estimation.

\textbf{ESCI}~\citep{reddy2022shopping}: released by Amazon as part of KDD Cup 2022, ESCI contains difficult search queries across US, Japan, and Spain marketplaces, each paired with up to 40 potentially relevant products. Each query-product pair is annotated with four relevance categories: Exact, Substitute, Complement, and Irrelevant. For our experiments, we preprocess ESCI by: (i) focusing on the US marketplace data; (ii) binarizing relevance judgments by considering only ``Exact'' and ``Irrelevant'' labels while dropping ambiguous ``Substitute'' and ``Complement'' cases; and (iii) selecting top-K ranked results and filtering queries with fewer than K results.

\textbf{Application data}: as our LLM-based query reformulation primarily affects the Body queries, we sample 8.5k of these and retrieve top-4 results from the production search system. We split this into 100 human-annotated queries and 8.4k unlabeled queries (84× labelled set size), providing a realistic scenario for applying \ours{} to production systems. We anonymize the data so that all identifiable user attributes were removed.

For the underlying search systems being evaluated, ESCI experiments use the dataset's inherent ranking, while our application uses a hybrid of boosted BM25-based lexical search and bi-encoder based semantic search.

\subsection{Automated Annotator models $M$}

For automated relevance judgment, we employ three models: (1) Claude 3 Sonnet and (2) Claude 3 Haiku with custom prompts incorporating uncertainty estimation, and (3) jina-reranker-v1-turbo-en, an off-the-shelf cross-encoder model. These models serve as synthetic annotators, providing relevance scores and confidence estimates for \ours{} calculations. 

For the LLM-based annotators, we prompted the model to elicit uncertainty levels (``About Even'', ``Slightly Better than Even'', ``Probably'', ``Pretty Good Chance'', ``Highly Likely'', ``Almost Certain'') which are mapped to numerical scores in [0.5, 1.0], with irrelevant predictions subtracting the mapped score from 1.0 (detailed prompts are in the Appendix). We apply isotonic regression calibration on the labelled set to improve score reliability. 

For evaluation, we compare two approaches: \textbf{prob} uses average annotator probability scores across K ranks as the Precision@K estimate, while \textbf{bin} binarizes scores using a 0.5 threshold before calculating Precision@K against the K-hot prediction vector. 

%% file: figure/esci_precision_results.tex
\begin{figure*}[t!]
    \centering
    \includegraphics[width=0.9\textwidth]{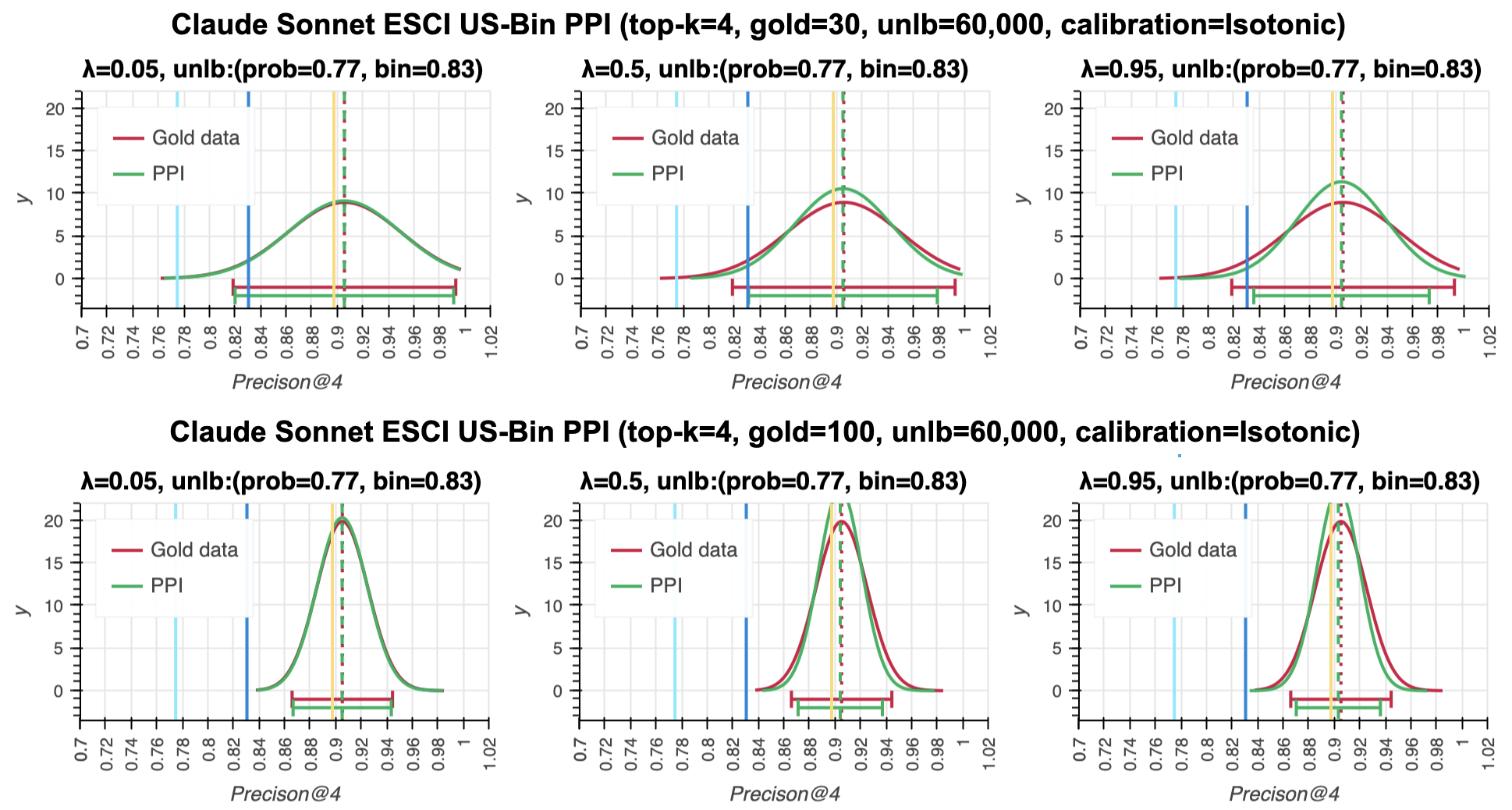}
    \caption{Estimated Precision@4 on ESCI. We show sampling distributions and 95\% CI for different estimators, calculated by sampling 50 gold datasets from ESCI. We consider samples of size $n=30$ (top row) and $n=100$, using $N=60,000$ unlabeled queries. Claude 3 Sonnet is used as the calibrated annotation model.
    The vertical yellow line denotes the true relevance, averaged across the entire ESCI dataset. \ours-PPI estimator (green) achieves variance reduction compared to the estimator using only Gold data (red), with superior reduction at higher $\lambda$ values. Both these approaches are significantly less biased than the LLM-only annotators \textbf{prob} (cyan) and \textbf{bin} (cerulean).
    }
    \label{fig:esci_precision_results}
\end{figure*}

%% file: sections/4b_Results.tex
\section{Analysis of \ours-PPI Estimator}

We first demonstrate the effectiveness of \ours{} by using the public ESCI dataset to analyse the correctness of our approach under controlled conditions where ground-truths for the unlabelled set are known.  

\subsubsection{Variance Reduction with Small Labelled Sets}

A key finding is that the \ours-PPI estimator provides substantial variance reduction even with as few as $n=30$ gold annotations. Figure~\ref{fig:esci_precision_results} shows the sampling distributions for Precision@4 estimation using different estimators; we observe that our approach demonstrates significantly tighter confidence intervals compared to gold-only estimates (red curves), indicating more reliable performance estimates. We also observe that LLM-only estimates are significantly biased for both \textbf{prob} and \textbf{bin} approaches.

\subsubsection{Optimal Unlabeled Set Size}

Our analysis reveals that large unlabeled sets are not necessary for effective estimation using \ours. Table~\ref{tab:unlabeled_size_analysis} shows the cost-performance trade-offs for different unlabeled set sizes. With $n=30$ gold samples, using 100x unlabeled data (3,000 unlabelled queries) provides nearly identical performance to using 2000x unlabeled data (60,000 queries), while reducing costs by 95\%. This finding is crucial for practical deployment, as it significantly reduces the computational cost of our approach.

\input{table/unlabeled_size_analysis}

\subsubsection{Cost-Performance Frontier}
Table~\ref{tab:esci_cost_performance} summarizes the cost-performance trade-offs for different annotator models. \mbox{Claude 3} Sonnet achieves the best bias-variance trade-off with a bias of only 0.70 points and standard error of 3.50, improving on gold-only estimation for the $n=30$ case. Notably, Claude 3 Haiku provides competitive performance at significantly lower cost (\$79 vs \$946 for Sonnet).

\input{table/precision_results}

\input{table/esci_cost_performance}

\subsubsection{LLM Judge Calibration}

We find that LLM-based evaluators (Claude 3 Sonnet and Haiku) demonstrate well-calibrated behavior, with most true positives receiving scores $\geq 0.5$ and true negatives receiving scores $\leq 0.4$. In contrast, the cross-encoder model (Jina Turbo) shows poor calibration with many true positives receiving low scores. Detailed calibration analysis is provided in the Appendix.

The calibration quality directly impacts PPI performance: calibrated models provide better variance reduction and more accurate estimates. This suggests that prompt-based uncertainty elicitation in LLMs is more effective than using off-the-shelf cross-encoder confidence scores for PPI applications.

The calibration analysis (in Appendix) provides additional insights into evaluator selection and the importance of uncertainty quantification for effective PPI implementation.

%% file: table/unlabeled_size_analysis.tex
\begin{table}[t!]
\centering
\small
\setlength{\tabcolsep}{3pt}
\begin{tabular}{llrrh}
\toprule
\textbf{Estimator} & \textbf{Unlb. Size} & \textbf{Bias \lowerbetter} & \textbf{Std. Error \lowerbetter} & \textbf{Cost (USD)} \\
\midrule
Gold & - & 1.04 & 4.45 & - \\
\midrule

Sonnet & \multirow{2}{*}{300 (10x)} & 1.04 & 3.45 & 4.73 \\
Haiku & & 1.07 & 4.02 & 0.40 \\
[0.25em]
Sonnet & \multirow{2}{*}{3k {} (100x)} & 0.52 & 3.67 & 47.28 \\
Haiku & & 0.42 & 4.10 & 3.97 \\
[0.25em]
Sonnet & \multirow{2}{*}{60k (2000x)} & 0.82 & 4.45 & 945.62 \\
Haiku & & 0.01 & 4.80 & 79.34 \\
\bottomrule
\end{tabular}
\caption{Effect of unlabeled set size on \ours-PPI estimator performance with $n=30$ gold samples. True Precision@4 = 89.73\%. The 100x configuration provides optimal trade-off.}
\label{tab:unlabeled_size_analysis}
\end{table}

%% file: table/precision_results.tex
\begin{table*}[t!]
\centering
\setlength{\tabcolsep}{4pt}
\begin{tabular}{lccccccccc}
\toprule
\multirow{2}{*}{\textbf{Estimator}} & \multicolumn{3}{c}{\textbf{Production Search}} & \multicolumn{3}{c}{\textbf{Reformulation V1}} & \multicolumn{3}{c}{\textbf{Reformulation V2}} \\
& K=1 & K=2 & K=4 & K=1 & K=2 & K=4 & K=1 & K=2 & K=4 \\
[0.5em]
\midrule

\multicolumn{10}{c}{
\underline{\textsc{Strict Relevance (Partial = Irrelevant)}}
} \\
[0.25em]
Gold ($n$=100) & 60.60\% & 60.00\% & 61.10\% & 64.60\% & 64.10\% & 65.70\% & 62.60\% & 62.60\% & 63.60\% \\
Sonnet-Unlb (prob) & 74.90\% & 74.90\% & 74.90\% & 77.40\% & 77.40\% & 77.40\% & 77.60\% & 77.60\% & 77.60\% \\
Sonnet-Unlb (binary) & 83.10\% & 82.70\% & 82.00\% & 85.20\% & 84.50\% & 84.00\% & 85.50\% & 84.80\% & 84.30\% \\
\ours-PPI ($\lambda$=0.95) & 55.10\% & 54.60\% & 55.30\% & 59.50\% & 59.10\% & 60.30\% & 59.70\% & 59.20\% & 59.40\% \\
[0.5em]
\multicolumn{10}{c}{
\underline{\textsc{Loose Relevance (Partial = Relevant)}}
} \\
[0.25em]
Gold ($n$=100) & 94.20\% & 93.70\% & 93.00\% & 97.30\% & 97.80\% & 97.60\% & 96.30\% & 97.30\% & 97.30\% \\
Sonnet-Unlb (prob) & 74.90\% & 74.90\% & 74.90\% & 77.40\% & 77.40\% & 77.40\% & 77.60\% & 77.60\% & 77.60\% \\
Sonnet-Unlb (binary) & 83.10\% & 82.70\% & 82.00\% & 85.20\% & 84.50\% & 84.00\% & 85.50\% & 84.80\% & 84.30\% \\
\ours-PPI ($\lambda$=0.95) & 91.40\% & 90.40\% & 89.50\% & 94.30\% & 94.50\% & 94.20\% & 94.00\% & 94.10\% & 94.30\% \\
\bottomrule
\end{tabular}
\caption{Precision@K Offline Metric Estimation for Query Reformulation. \textbf{Note:} we anonymize these numbers by introducing a randomly-selected value as the baseline.}
\label{tab:application_precision}
\vspace{1ex}
\end{table*}

%% file: table/esci_cost_performance.tex
\begin{table}[t]
\centering
\small
\begin{tabular}{lrrr}
\toprule
\textbf{Estimator} & \textbf{Bias \lowerbetter} & \textbf{Std. Error \lowerbetter} & \textbf{Cost (USD)} \\
\midrule
Gold & 1.04 & 4.45 & - \\
\midrule
Claude 3 Sonnet & 0.70 & 3.50 & 945.6 \\
Claude 3 Haiku & 0.29 & 3.86 & 79.3 \\
Jina Turbo & 0.51 & 4.26 & $<$5.0 \\
\bottomrule
\end{tabular}
\caption{Cost-performance comparison for Precision@4 estimation on ESCI with $N=60,000$ unlabeled queries and $n=30$ gold samples. We measure Bias and Std. Error of the estimator as performance metrics.}
\label{tab:esci_cost_performance}
\end{table}

%% file: sections/4c_Results_Application.tex
\section{Production Deployment Results}
\label{sec:application_results}

Having validated \ours{} on the ESCI dataset, we demonstrate its real-world applicability by deploying an LLM-based query reformulation system in the production \mbox{e-commerce} search application and measuring its impact. We mention the evaluation prompt in the Appendix.

\subsubsection{Query Reformulation System Design}

We developed two LLM-based query reformulation treatments using Claude 3 Sonnet with few-shot Chain-of-Thought prompting to address the query defects identified in our analysis:

\begin{itemize}
    \item \textbf{Treatment 1 (T1)}: Basic query reformulation performing Hinglish-to-English translation and correction of grammatical errors and typos (V1 Prompt).
    \item \textbf{Treatment 2 (T2)}: Enhanced reformulation with Indian ethnic context preservation (e.g., retaining "kurti", "salwar kameez") (V2 Prompt) plus rule-based word-level correction for cache misses.
\end{itemize}

Both treatments target Head and Body queries (covering 75\% of search volume), while excluding Tail queries due to their uniqueness (the vast majority are searched only once) and prohibitive size (several million queries). The system processes queries through a reformulation cache for fast realtime processing.

\subsubsection{PPI-Based Pre-Deployment Evaluation}

Prior to production deployment, we applied \ours{} to estimate Precision@K improvements across treatments. Using our instance-level formulation on 8,500 Body queries (n=100 gold, N=8,400 unlabeled with 84× ratio), we obtained rapid evaluation results within 2 hours of human annotation by domain experts.

Table~\ref{tab:application_precision} shows the Precision@K estimates of various approaches. Under strict relevance criteria, T1 demonstrates clear improvements over the control (C) across all K values (+13.4\% relative improvement in Precision@4). T2 shows similar but slightly lower gains. Notably, our PPI estimates predicted T1 would outperform T2, which was later confirmed in production deployment.

This offline analysis using \ours{} provided crucial confidence for deployment decisions, demonstrating that our method accurately estimates true performance improvements even when the relative differences between treatments are subtle.

\subsubsection{Production A/B Test Results}

We conducted an A/B experiment comparing Control (C), Treatment 1 (T1), and Treatment 2 (T2) across the entire application. The A/B test results validated estimates from our method and demonstrated significant business impact from the T1 treatment, which was finally deployed.

\paragraph{Business Impact Validation}
The production deployment results presented in Table~\ref{tab:application_business} demonstrate significant business impact across all key application metrics, validating our \ours-based estimates: T1 achieved superior performance compared to both the control and T2, exhibiting a 407bps improvement in daily business-as-usual sales. Notably, customer purchasing behavior improved with a 90bps increase in orders per customer, while the average selling price increased by 137bps, indicating that customers were successfully discovering higher-value products through improved query reformulation. Treatment 2 showed positive but comparatively weaker improvements with a 174bps increase in daily sales, confirming our method's ability to accurately predict relative treatment preference.

\input{table/business_metrics}

\paragraph{Search Quality Improvements}
The query-level analysis presented in Table~\ref{tab:application_search} reveals consistent improvements in search quality metrics for Treatment 1 across all measured dimensions. Most notably, T1 achieved a 571bps improvement in click-through rates for reformulated queries, accompanied by a 304bps increase in clicks per query session, indicating enhanced user engagement with search results. 

Perhaps most significantly, the results demonstrate improved search engagement: customers browsed 7.82\% deeper into search pages and clicked more per query session, suggesting that reformulated queries better captured user intent and reduced the need for query refinement. For Hinglish queries, T1 demonstrated particularly strong performance with 579bps improvement in browsing depth and 494bps increase in clicks per customer, validating the effectiveness of our Hinglish-to-English translation approach. 

These improvements are particularly noteworthy given that T1 represents a relatively simple reformulation strategy compared to the more sophisticated T2 treatment, highlighting the counterintuitive finding that basic translation and error correction can outperform more complex contextual preservation approaches.

\input{table/search_metrics}

\subsubsection{Economic Impact and Scalability}

The deployment demonstrated exceptional economic viability with significant return on investment. The implementation required a one-time reformulation cost for millions of Head and Body queries, resulting in substantial annualized revenue improvements that yielded a several-fold return on investment. Examples of the query reformulations produced by each treatment are provided in the Appendix.

The deployment success has enabled expansion to additional query types and search improvements, demonstrating the practical scalability of PPI-guided ML deployment in real-world e-commerce environments.

%% file: table/business_metrics.tex
\begin{table}[t!]
\centering
\begin{tabular}{lrr}
\toprule
\textbf{Metric} & \textbf{T1} & \textbf{T2} \\
\midrule
Avg. orders per customer & \textbf{+90 bps} & +42 bps \\
Avg. add-to-cart per customer & \textbf{+6 bps} & +5 bps \\
BAU Daily Sales & \textbf{+407 bps} & +174 bps \\
Avg. Sale Price & \textbf{+137 bps} & +11 bps \\
\bottomrule
\end{tabular}
\caption{
Application-level business impact metrics from an equal-allocation A/B test comparing two query reformulation approaches. Treatment 1 (T1) applies query reformulation on Head and Body queries, while Treatment 2 (T2) uses rule-based correction. Results show improvements in basis points (bps) across key business indicators, with T1 consistently outperforming T2 (bolded values indicate best performance).
}
\label{tab:application_business}
\end{table}

%% file: table/search_metrics.tex
\begin{table}[t!]
\centering
{%
\begin{tabular}{lrr}
\toprule
\textbf{Metric} & \textbf{T1} & \textbf{T2} \\
\midrule

\multicolumn{3}{c}{\underline{\textsc{All Corrected Queries}}} \\
[0.25em]
(CTR) Click-through rate & \textbf{+571 bps} & +426 bps \\
(CPQ) Clicks per query session & \textbf{+304 bps} & +93 bps \\
(CPC) Clicks per customer & \textbf{+404 bps} & +174 bps \\
Avg. Browse Depth & \textbf{+782 bps} & +614 bps \\
[0.25em]
\multicolumn{3}{c}{\underline{\textsc{Hinglish Queries}}} \\
[0.25em]
(CTR) Click-through rate & \textbf{+77 bps} & -154 bps \\
(CPQ) Clicks per query session & \textbf{+406 bps} & -259 bps \\
(CPC) Clicks per customer & \textbf{+494 bps} & -233 bps \\
Avg. Browse Depth & \textbf{+579 bps} & +214 bps \\
\bottomrule
\end{tabular} %
}
\footnotesize
\begin{tabular}{l}
\end{tabular}
\caption{
Search quality metrics comparing two treatments (T1 and T2) against baseline. Results show improvements in basis points (bps) across key search experience indicators. T1 consistently outperforms T2 across all metrics (bolded values indicate best performance). For Hinglish queries specifically, T1 shows positive gains while T2 shows negative impact on several metrics.
}
\label{tab:application_search}
\end{table}

%% file: sections/5_Conclusion.tex
\section{Lessons Learned During Development, Deployment, and Maintenance}

Throughout the development and deployment process, several lessons were learned:

\begin{enumerate}
    \item \textbf{\ours{} enables rapid deployment decisions.} The A/B test demonstrated that \ours-PPI based estimation can be completed in 2 hours of domain expert annotation versus weeks for traditional approaches. Our offline estimates correctly predicted treatment preference (T1 $>$ T2 $>$ Control) and relative performance magnitudes, which were subsequently validated in production A/B testing.

    \item \textbf{Cultural context preservation requires domain expertise.} Treatment 2's enhanced prompting with Indian ethnic context (preserving terms like "kurti", "salwar kameez") initially appeared superior in offline analysis but was outperformed by simpler Treatment 1 in production. This counterintuitive finding suggests that basic translation and error correction can be more effective than complex contextual preservation, highlighting the importance of A/B testing to validate \ours-guided decisions.

    \item \textbf{\ours{} plateaus with unlabeled data size.} Increasing unlabeled data from 10x to 2000x the gold set size showed diminishing returns. With $n$=30 gold samples, using 100x unlabeled data (3,000 queries) provided nearly identical performance to 2000x unlabeled data (60,000 queries) while reducing costs by 95\%. This suggests that investing in more gold data is more beneficial than scaling unlabeled data beyond 100x.

    \item \textbf{Calibration is critical for LLM-based judges.} Our experiments showed that calibrated relevance scores using isotonic regression consistently outperformed uncalibrated scores across all judge models. Even with as few as 30 gold datapoints, calibration provided better PPI estimates with lower variance. LLM-based evaluators (Claude 3 Sonnet/Haiku) demonstrated well-calibrated behavior with most true positives receiving scores $\geq 0.5$, while cross-encoder models (Jina Turbo) showed poor calibration with many true positives receiving low scores $\leq 0.4$.
    
    \item \textbf{Model choice significantly impacts cost-performance tradeoffs.} Claude 3 Haiku achieved comparable performance to Sonnet (bias: 0.29 vs 0.70, standard error: 3.86 vs 3.50) at 12x lower cost (\$79 vs \$946 for 60k queries). Off-the-shelf cross-encoder models showed poor calibration and barely improved variance compared to gold-only estimation, making prompt-based uncertainty elicitation in LLMs more effective than cross-encoder confidence scores for PPI applications.

\end{enumerate}
\input{figure/calibration_comparison}

\section{Conclusion}
We presented PRECISE, a statistical framework that significantly reduces the human annotation burden in evaluating ranking systems by combining minimal human judgments with LLM-based assessments. 

Our approach achieves reliable metric estimation using as few as 100 human-annotated queries while correcting for inherent LLM biases. Through our novel formulation using sparse K-hot vectors and rank-level decomposition, we made prediction-powered inference computationally tractable for large-scale ranking evaluation.

The success of PRECISE opens up new possibilities for efficient, scalable evaluation of information retrieval systems while maintaining high confidence in the resulting metrics. As LLM capabilities continue to advance, we expect frameworks like PRECISE (and more generally, PPI-style estimation) to become increasingly valuable in both research and production environments.

%% file: figure/calibration_comparison.tex
\begin{figure*}[t!]
    \centering
    \includegraphics[width=0.95\textwidth]{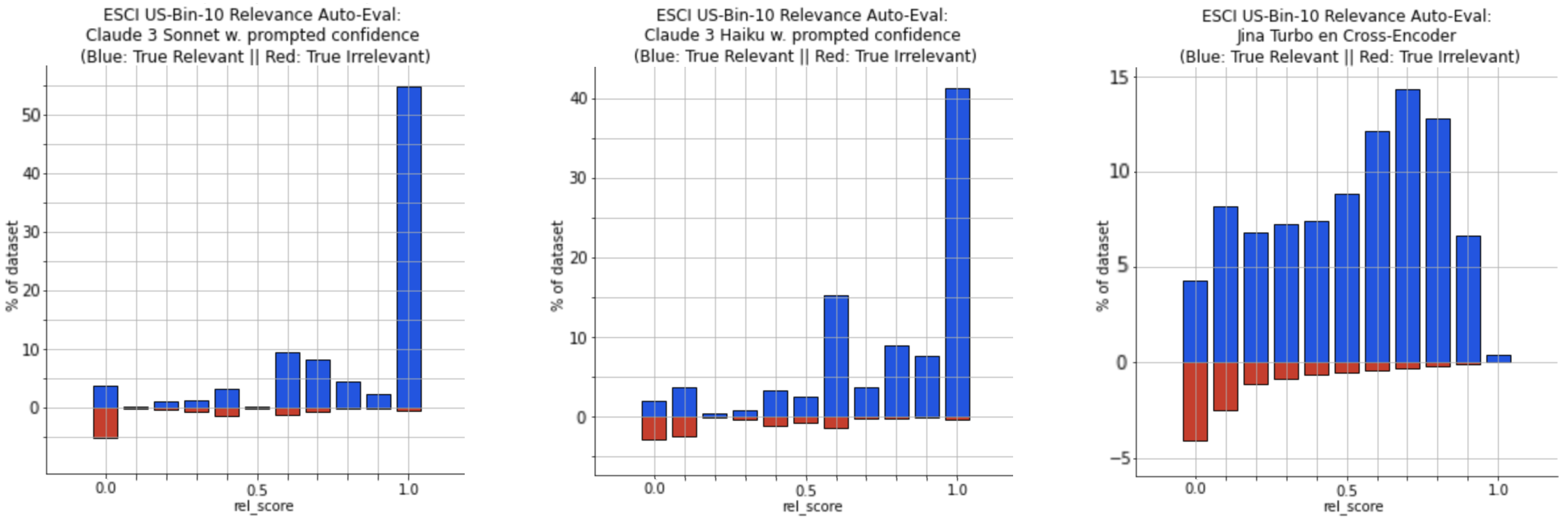}
    \caption{Calibration comparison across LLM evaluator models. Left: Claude 3 Sonnet (well-calibrated), Center: Claude 3 Haiku (moderately calibrated), Right: Jina Turbo (poorly calibrated). Blue bars represent true positives, red bars represent true negatives.}
    \label{fig:calibration_comparison}
\end{figure*}

%% file: sections/6_Future_work.tex
\section{Future Work}
Several promising directions remain for future work. We describe a few of them below:

\begin{enumerate}
    \item The reliance on a ``gold'' (human-labelled) set is the major bottlenecks of any estimation method. Instead, LLM-generated synthetic datasets can provide ``silver'' labels which may still be usable for estimation \cite{kowshik-etal-2024-corrsynth, divekar-durrett-2024-synthesizrr}.
    
    \item Extending \ours{} to handle dynamic corpus updates, where new documents are continuously added to the retrieval system, would enhance its practical utility in production environments. Recent approaches in generative retrieval over evolving corpora \cite{zhang2024generative} highlight the need for statistically robust metrics that can adapt without full re-annotation.

    \item Multi-turn conversational search and multi-modal retrieval provide an alternate scope for investigating the framework's applicability to sub-example level estimates. Evaluating these complex modalities often requires intricate user simulation or comprehensive multi-modal benchmarks \cite{Fu2023MMEAC}, presenting unique challenges for bias correction in metric estimation.

    \item Another promising direction involves developing methods to combine judgments from multiple LLMs with different strengths and biases. Ensembling LLM judges has been shown to align better with human preferences than single-model evaluators \cite{zheng2024judging}, potentially leading to more robust assessments within the \ours{} framework.

    \item Finally, adapting the framework for online evaluation settings where relevance assessments need to be generated in real-time would broaden its applicability. Doubly robust estimation for online ranking \cite{oosterhuis2023doubly} shares theoretical grounds with LLM bias and could offer a pathway toward real-time, bias-corrected metric inference.
\end{enumerate}

%% file: appendices/calibration.tex
\section{LLM-as-a-Judge Calibration Analysis}
\label{app:calibration}

Here, we provide a detailed analysis of the calibration properties of different LLM judge models used in our experiments.

\subsection{Calibration Methodology}

We evaluate calibration by examining the distribution of confidence scores assigned by each judge model to true positive (actually relevant) and true negative (actually irrelevant) query-document pairs. In an ideally calibrated system, all actually relevant pairs should receive scores close to 1.0, while irrelevant pairs should receive scores close to 0.0.

\subsection{LLM Judges}

\subsubsection{Claude 3 Sonnet}
Claude 3 Sonnet demonstrates excellent calibration behavior. Nearly all true positives receive scores $\geq 0.5$, with the majority concentrated at higher confidence levels (0.8-1.0). True negatives are well-separated, with most receiving scores $\leq 0.4$. This clear separation between relevant and irrelevant items contributes to the model's effectiveness in PPI estimation.

\subsubsection{Claude 3 Haiku}
Claude 3 Haiku shows slightly weaker calibration compared to Sonnet, with some true positives receiving lower scores (0.6-0.8 range). However, the overall calibration is still reasonable, with most true positives above 0.5 and most true negatives below 0.4. The reduced calibration quality compared to Sonnet may explain its slightly higher standard error in PPI estimation.

\subsection{Cross-Encoder Model}

\subsubsection{Jina-reranker-v1-turbo-en}
The Jina Turbo cross-encoder shows poor calibration, with a high proportion of true positives receiving scores $\leq 0.4$. While true negatives are well-calibrated (correctly receiving low scores), the systematic underestimation of relevance for actually relevant pairs severely impacts the model's utility for PPI. This poor calibration explains why Jina Turbo barely improves variance compared to gold-only estimation.

\subsection{Impact on PPI Performance}

The calibration quality directly correlates with PPI effectiveness:

\begin{itemize}
    \item \textbf{Better-calibrated models} (Claude 3 Sonnet, Haiku) provide substantial variance reduction and accurate bias correction
    \item \textbf{Poorly calibrated models} (Jina Turbo) offer minimal improvement over gold-only estimation
    \item \textbf{Calibration correction} using isotonic regression on the gold set improves performance for all models, but the improvement is most pronounced for poorly calibrated models
\end{itemize}

\subsection{Recommendations}

Based on our calibration analysis, we recommend:

\begin{enumerate}
    \item \textbf{Prefer LLM judges} with prompted uncertainty over off-the-shelf cross-encoder models
    \item \textbf{Apply calibration correction} (e.g., isotonic regression) when possible, especially for weaker models
    \item \textbf{Evaluate calibration quality} before deploying any LLM judge in a PPI-style framework
    \item \textbf{Consider cost-performance trade-offs}: Claude 3 Haiku provides good calibration at significantly lower cost than Sonnet
\end{enumerate}

%% file: appendices/query_prompts.tex
\begin{table}[th]
\centering
{\scriptsize\ttfamily
\begin{tabular}{p{0.42\textwidth}}
\toprule
{\small \textnormal{\textbf{LLM-as-a-Judge Relevance Annotation Prompt}}}
 \\
\midrule
<role> \\
You are an expert product judge who works for e-commerce website Amazon. Your job is to determine if a particular product is relevant to a search query asked by Amazon customers. This is to improve the experience and safety of the customers. Make sure you output XML when asked. \\
</role> \\
<task> \\
The customer's search query is mentioned in <search-query></search-query> XML tags. The product details are mentioned in <product-details></product-details> XML tags. \\
1. First, output your thoughts in <thinking></thinking> XML tags. Here, enter your justification and reasoning for your evaluation. \\
2. Secondly, output your evaluation of the relevance of the product to the search query. Your evaluation of the response should be output in <evaluation></evaluation> XML tags. Conduct your evaluation of the relevance between the search query and product as follows: \\
- Relevant: If the product details exactly or partially relates to the search query, output <evaluation>Relevant</evaluation>. Consider partial matches which fulfill some but not all criterion in the search query, should be considered Relevant. \\
- Irrelevant: If the product details does not have any match to the search query, output <evaluation>Irrelevant</evaluation>. Unrelated products and complementary products which do not match the search query, should be considered Irrelevant. \\
3. Finally, provide your best guess for how confident you are that your evaluation is correct in <confidence></confidence> XML tags. Give ONLY your confidence, no other words or explanation. Provide your confidence label as exactly following expressions (ordered from least confident to most confident): \\
- About Even \\
- Slightly Better than Even \\
- Probably \\
- Pretty Good Chance \\
- Highly Likely \\
- Almost Certain \\
</task> \\
\hline
\end{tabular}
}
\caption{LLM-as-a-Judge Relevance Annotation Prompt}
\label{tab:relevance_prompt}
\end{table}

\section{Relevance Annotation Prompt}
\label{app:relevance_prompt}
Table~\ref{tab:relevance_prompt} presents the relevance judge prompt used in our production deployment.

%% file: aaai2026.bib
@article{reddy2022shopping,
title={Shopping Queries Dataset: A Large-Scale {ESCI} Benchmark for Improving Product Search},
author={Chandan K. Reddy and Lluís Màrquez and Fran Valero and Nikhil Rao and Hugo Zaragoza and Sambaran Bandyopadhyay and Arnab Biswas and Anlu Xing and Karthik Subbian},
year={2022},
eprint={2206.06588},
archivePrefix={arXiv}
}

@inproceedings{boyeau2025autoeval,
    title={AutoEval Done Right: Using Synthetic Data for Model Evaluation},
    author={Pierre Boyeau and Anastasios Nikolas Angelopoulos and Tianle Li and Nir Yosef and Jitendra Malik and Michael I. Jordan},
    booktitle={Forty-second International Conference on Machine Learning},
    year={2025},
    url={https://openreview.net/forum?id=S8kbmk12Oo}
}

@misc{angelopoulos2024ppiefficientpredictionpoweredinference,
      title={PPI++: Efficient Prediction-Powered Inference}, 
      author={Anastasios N. Angelopoulos and John C. Duchi and Tijana Zrnic},
      year={2024},
      eprint={2311.01453},
      archivePrefix={arXiv},
      primaryClass={stat.ML},
      url={https://arxiv.org/abs/2311.01453}, 
}

@article{
doi:10.1126/science.adi6000,
author = {Anastasios N. Angelopoulos  and Stephen Bates  and Clara Fannjiang  and Michael I. Jordan  and Tijana Zrnic },
title = {Prediction-powered inference},
journal = {Science},
volume = {382},
number = {6671},
pages = {669-674},
year = {2023},
doi = {10.1126/science.adi6000},
URL = {https://www.science.org/doi/abs/10.1126/science.adi6000},
eprint = {https://www.science.org/doi/pdf/10.1126/science.adi6000}}

@inproceedings{shen-etal-2023-large,
    title = "Large Language Models are Not Yet Human-Level Evaluators for Abstractive Summarization",
    author = "Shen, Chenhui  and
      Cheng, Liying  and
      Nguyen, Xuan-Phi  and
      You, Yang  and
      Bing, Lidong",
    editor = "Bouamor, Houda  and
      Pino, Juan  and
      Bali, Kalika",
    booktitle = "Findings of the Association for Computational Linguistics: EMNLP 2023",
    month = dec,
    year = "2023",
    address = "Singapore",
    publisher = "Association for Computational Linguistics",
    url = "https://aclanthology.org/2023.findings-emnlp.278/",
    doi = "10.18653/v1/2023.findings-emnlp.278",
    pages = "4215--4233"
}

@inproceedings{li-etal-2024-split,
    title = "Split and Merge: Aligning Position Biases in {LLM}-based Evaluators",
    author = "Li, Zongjie  and
      Wang, Chaozheng  and
      Ma, Pingchuan  and
      Wu, Daoyuan  and
      Wang, Shuai  and
      Gao, Cuiyun  and
      Liu, Yang",
    editor = "Al-Onaizan, Yaser  and
      Bansal, Mohit  and
      Chen, Yun-Nung",
    booktitle = "Proceedings of the 2024 Conference on Empirical Methods in Natural Language Processing",
    month = nov,
    year = "2024",
    address = "Miami, Florida, USA",
    publisher = "Association for Computational Linguistics",
    url = "https://aclanthology.org/2024.emnlp-main.621/",
    doi = "10.18653/v1/2024.emnlp-main.621",
    pages = "11084--11108"
}

@inproceedings{10.1145/2911451.2911537,
author = {Wang, Xuanhui and Bendersky, Michael and Metzler, Donald and Najork, Marc},
title = {Learning to Rank with Selection Bias in Personal Search},
year = {2016},
isbn = {9781450340694},
publisher = {Association for Computing Machinery},
address = {New York, NY, USA},
url = {https://doi.org/10.1145/2911451.2911537},
doi = {10.1145/2911451.2911537},
booktitle = {Proceedings of the 39th International ACM SIGIR Conference on Research and Development in Information Retrieval},
pages = {115–124},
numpages = {10},
location = {Pisa, Italy},
series = {SIGIR '16}
}

@inproceedings{10.1145/3366423.3380255,
author = {Ovaisi, Zohreh and Ahsan, Ragib and Zhang, Yifan and Vasilaky, Kathryn and Zheleva, Elena},
title = {Correcting for Selection Bias in Learning-to-rank Systems},
year = {2020},
isbn = {9781450370233},
publisher = {Association for Computing Machinery},
address = {New York, NY, USA},
url = {https://doi.org/10.1145/3366423.3380255},
doi = {10.1145/3366423.3380255},
booktitle = {Proceedings of The Web Conference 2020},
pages = {1863–1873},
numpages = {11},
location = {Taipei, Taiwan},
series = {WWW '20}
}

@inproceedings{es-etal-2024-ragas,
    title = "{RAGA}s: Automated Evaluation of Retrieval Augmented Generation",
    author = "Es, Shahul  and
      James, Jithin  and
      Espinosa Anke, Luis  and
      Schockaert, Steven",
    editor = "Aletras, Nikolaos  and
      De Clercq, Orphee",
    booktitle = "Proceedings of the 18th Conference of the European Chapter of the Association for Computational Linguistics: System Demonstrations",
    month = mar,
    year = "2024",
    address = "St. Julians, Malta",
    publisher = "Association for Computational Linguistics",
    url = "https://aclanthology.org/2024.eacl-demo.16/",
    pages = "150--158"
}

@inproceedings{saad-falcon-etal-2024-ares,
    title = "{ARES}: An Automated Evaluation Framework for Retrieval-Augmented Generation Systems",
    author = "Saad-Falcon, Jon  and
      Khattab, Omar  and
      Potts, Christopher  and
      Zaharia, Matei",
    editor = "Duh, Kevin  and
      Gomez, Helena  and
      Bethard, Steven",
    booktitle = "Proceedings of the 2024 Conference of the North American Chapter of the Association for Computational Linguistics: Human Language Technologies (Volume 1: Long Papers)",
    month = jun,
    year = "2024",
    address = "Mexico City, Mexico",
    publisher = "Association for Computational Linguistics",
    url = "https://aclanthology.org/2024.naacl-long.20/",
    doi = "10.18653/v1/2024.naacl-long.20",
    pages = "338--354"
}

@inproceedings{10.5555/3666122.3668142,
author = {Zheng, Lianmin and Chiang, Wei-Lin and Sheng, Ying and Zhuang, Siyuan and Wu, Zhanghao and Zhuang, Yonghao and Lin, Zi and Li, Zhuohan and Li, Dacheng and Xing, Eric P. and Zhang, Hao and Gonzalez, Joseph E. and Stoica, Ion},
title = {Judging LLM-as-a-judge with MT-bench and Chatbot Arena},
year = {2023},
publisher = {Curran Associates Inc.},
address = {Red Hook, NY, USA},
booktitle = {Proceedings of the 37th International Conference on Neural Information Processing Systems},
articleno = {2020},
numpages = {29},
location = {New Orleans, LA, USA},
series = {NIPS '23}
}

@inproceedings{dong-etal-2024-llm,
    title = "Can {LLM} be a Personalized Judge?",
    author = "Dong, Yijiang River  and
      Hu, Tiancheng  and
      Collier, Nigel",
    editor = "Al-Onaizan, Yaser  and
      Bansal, Mohit  and
      Chen, Yun-Nung",
    booktitle = "Findings of the Association for Computational Linguistics: EMNLP 2024",
    month = nov,
    year = "2024",
    address = "Miami, Florida, USA",
    publisher = "Association for Computational Linguistics",
    url = "https://aclanthology.org/2024.findings-emnlp.592/",
    doi = "10.18653/v1/2024.findings-emnlp.592",
    pages = "10126--10141"
}

@inproceedings{chen-etal-2024-humans,
    title = "Humans or {LLM}s as the Judge? A Study on Judgement Bias",
    author = "Chen, Guiming Hardy  and
      Chen, Shunian  and
      Liu, Ziche  and
      Jiang, Feng  and
      Wang, Benyou",
    editor = "Al-Onaizan, Yaser  and
      Bansal, Mohit  and
      Chen, Yun-Nung",
    booktitle = "Proceedings of the 2024 Conference on Empirical Methods in Natural Language Processing",
    month = nov,
    year = "2024",
    address = "Miami, Florida, USA",
    publisher = "Association for Computational Linguistics",
    url = "https://aclanthology.org/2024.emnlp-main.474/",
    doi = "10.18653/v1/2024.emnlp-main.474",
    pages = "8301--8327"
}

@inproceedings{Achiam2023GPT4TRShort,
  title={GPT-4 Technical Report},
  author={OpenAI Josh Achiam and Steven Adler and others},
  year={2023},
  url={https://api.semanticscholar.org/CorpusID:257532815}
}

@misc{bai2022constitutionalaiharmlessnessaiShort,
      title={Constitutional AI: Harmlessness from AI Feedback}, 
      author={Yuntao Bai and Saurav Kadavath and others},
      year={2022},
      eprint={2212.08073},
      archivePrefix={arXiv},
      primaryClass={cs.CL},
      url={https://arxiv.org/abs/2212.08073}, 
}

@misc{deepseekai2025deepseekv3technicalreportShort,
      title={DeepSeek-V3 Technical Report}, 
      author={DeepSeek-AI and Aixin Liu and Bei Feng and others},
      year={2025},
      eprint={2412.19437},
      archivePrefix={arXiv},
      primaryClass={cs.CL},
      url={https://arxiv.org/abs/2412.19437}, 
}

@inproceedings{divekar-durrett-2024-synthesizrr,
    title = "{S}ynthesiz{RR}: Generating Diverse Datasets with Retrieval Augmentation",
    author = "Divekar, Abhishek  and
      Durrett, Greg",
    editor = "Al-Onaizan, Yaser  and
      Bansal, Mohit  and
      Chen, Yun-Nung",
    booktitle = "Proceedings of the 2024 Conference on Empirical Methods in Natural Language Processing",
    month = nov,
    year = "2024",
    address = "Miami, Florida, USA",
    publisher = "Association for Computational Linguistics",
    url = "https://aclanthology.org/2024.emnlp-main.1071/",
    doi = "10.18653/v1/2024.emnlp-main.1071",
    pages = "19200--19227",
}

@inproceedings{kowshik-etal-2024-corrsynth,
    title = "{C}orr{S}ynth - A Correlated Sampling Method for Diverse Dataset Generation from {LLM}s",
    author = "Kowshik, Suhas S  and
      Divekar, Abhishek  and
      Malik, Vijit",
    editor = "Al-Onaizan, Yaser  and
      Bansal, Mohit  and
      Chen, Yun-Nung",
    booktitle = "Proceedings of the 2024 Conference on Empirical Methods in Natural Language Processing",
    month = nov,
    year = "2024",
    address = "Miami, Florida, USA",
    publisher = "Association for Computational Linguistics",
    url = "https://aclanthology.org/2024.emnlp-main.899/",
    doi = "10.18653/v1/2024.emnlp-main.899",
    pages = "16076--16095",
}

@inproceedings{zhang2024generative,
  author = {Zhang, Zhen and Ma, Xinyu and Sun, Weiwei and Ren, Pengjie and Chen, Zhumin and Wang, Shuaiqiang and Yin, Dawei and de Rijke, Maarten and Ren, Zhaochun},
  title = {Replication and Exploration of Generative Retrieval over Dynamic Corpora},
  year = {2025},
  isbn = {9798400715921},
  publisher = {Association for Computing Machinery},
  address = {New York, NY, USA},
  url = {https://doi.org/10.1145/3726302.3730314},
  doi = {10.1145/3726302.3730314},
  booktitle = {Proceedings of the 48th International ACM SIGIR Conference on Research and Development in Information Retrieval},
  pages = {3325–3334},
  numpages = {10},
  location = {Padua, Italy},
  series = {SIGIR '25}
}

@article{Fu2023MMEAC,
  title={MME: A Comprehensive Evaluation Benchmark for Multimodal Large Language Models},
  author={Chaoyou Fu and Peixian Chen and Yunhang Shen and Yulei Qin and Mengdan Zhang and Xu Lin and Zhenyu Qiu and Wei Lin and Jinrui Yang and Xiawu Zheng and Ke Li and Xing Sun and Rongrong Ji},
  journal={ArXiv},
  year={2023},
  volume={abs/2306.13394},
  url={https://api.semanticscholar.org/CorpusID:259243928}
}

@inproceedings{zheng2024judging,
  author = {Zheng, Lianmin and Chiang, Wei-Lin and Sheng, Ying and Zhuang, Siyuan and Wu, Zhanghao and Zhuang, Yonghao and Lin, Zi and Li, Zhuohan and Li, Dacheng and Xing, Eric P. and Zhang, Hao and Gonzalez, Joseph E. and Stoica, Ion},
  title = {Judging LLM-as-a-judge with MT-bench and Chatbot Arena},
  year = {2023},
  publisher = {Curran Associates Inc.},
  address = {Red Hook, NY, USA},
  booktitle = {Proceedings of the 37th International Conference on Neural Information Processing Systems},
  articleno = {2020},
  numpages = {29},
  location = {New Orleans, LA, USA},
  series = {NIPS '23}
}

@article{oosterhuis2023doubly,
  author = {Oosterhuis, Harrie},
  title = {Doubly Robust Estimation for Correcting Position Bias in Click Feedback for Unbiased Learning to Rank},
  year = {2023},
  issue_date = {July 2023},
  publisher = {Association for Computing Machinery},
  address = {New York, NY, USA},
  volume = {41},
  number = {3},
  issn = {1046-8188},
  url = {https://doi.org/10.1145/3569453},
  doi = {10.1145/3569453},
  journal = {ACM Trans. Inf. Syst.},
  month = feb,
  articleno = {61},
  numpages = {33}
}
